\documentclass[conference]{IEEEtran}
\IEEEoverridecommandlockouts
\usepackage{cite}
\usepackage{amsmath,amssymb,amsfonts}
\usepackage{algorithmic}
\usepackage{graphicx}
\usepackage{textcomp}
\usepackage{xcolor}
\usepackage{hyperref}  
\usepackage{longtable,booktabs,multirow} 
\usepackage[ruled,vlined]{algorithm2e}
\usepackage{orcidlink}

\newcommand{\norm}[1]{\left\lVert#1\right\rVert} 
\DeclareMathOperator*{\argmin}{argmin}
\DeclareMathOperator*{\argmax}{argmax}

\newcommand{\supptitle}[1]{\def\mySuppTitleContent{#1}}
\newcommand{\makeMySupplementaryTitle}{%
  \twocolumn[{%
    \begin{center}
      \vspace*{1em} 
      {\LARGE\bfseries \mySuppTitleContent\par}
      \vspace{1.5em}
      {\large\itshape\textemdash\quad Supplementary Materials\quad\textemdash\par}
      \vspace{2em} 
    \end{center}
  }]%
}

\def\eg{\emph{e.g}.} \def\Eg{\emph{E.g}.}
\def\ie{\emph{i.e}.} \def\Ie{\emph{I.e}.}
\def\cf{\emph{c.f}.} \def\Cf{\emph{C.f}.}
\def\etc{\emph{etc}.} \def\vs{\emph{vs}.}
\def\wrt{w.r.t.} \def\dof{d.o.f.}
\def\etal{\emph{et al}.}

\renewcommand{\figureautorefname}{Fig.}
\renewcommand{\tableautorefname}{Table} 
\renewcommand{\equationautorefname}{Eqn.} 
\renewcommand{\sectionautorefname}{\S} 
\renewcommand{\subsectionautorefname}{\S} 

\def\BibTeX{{\rm B\kern-.05em{\sc i\kern-.025em b}\kern-.08em
    T\kern-.1667em\lower.7ex\hbox{E}\kern-.125emX}}
\begin{document}

\title{CAPAA: Classifier-Agnostic Projector-Based Adversarial Attack}
  
\author{
\IEEEauthorblockN{\textit{Zhan Li}~\orcidlink{0009-0007-8680-4729}$^{1,}$\IEEEauthorrefmark{1},
\textit{Mingyu Zhao}~\orcidlink{0009-0002-7386-9519}$^{1,2,}$\IEEEauthorrefmark{1}, 
\textit{Xin Dong}~\orcidlink{0009-0007-0540-0010}$^{1}$,
\textit{Haibin Ling}~\orcidlink{0000-0003-4094-8413}$^{3}$,
\textit{Bingyao Huang}~\orcidlink{0000-0002-8647-5730}$^{1,}$\IEEEauthorrefmark{2}
}
\IEEEauthorblockN{$^{1}$Southwest University, China \quad
$^{2}$Rutgers University, USA \quad
$^{3}$Stony Brook University, USA}

\thanks{\IEEEauthorrefmark{1}These authors contributed equally.}
    \thanks{Zhan Li and Xin Dong are with Southwest University. E-mail: \{lz20020722, dongxin12345\}@email.swu.edu.cn.}
    \thanks{Mingyu Zhao is with Rutgers University. Work partly done during internship with Southwest University. E-mail: mz751@scarletmail.rutgers.edu. }
    \thanks{Haibin Ling is with Dept. of Computer Science, Stony Brook University. E-mail: hling@cs.stonybrook.edu.}
    \thanks{\IEEEauthorrefmark{2}Bingyao Huang is the corresponding author. E-mail: bhuang@swu.edu.cn.}
}
\maketitle

\begin{abstract}

Projector-based adversarial attack aims to project carefully designed light patterns (i.e., adversarial projections) onto scenes to deceive deep image classifiers. It has potential applications in privacy protection and the development of more robust classifiers. However, existing approaches primarily focus on individual classifiers and fixed camera poses, often neglecting the complexities of multi-classifier systems and scenarios with varying camera poses. This limitation reduces their effectiveness when introducing new classifiers or camera poses. In this paper, we introduce Classifier-Agnostic Projector-Based Adversarial Attack (CAPAA) to address these issues. First, we develop a novel classifier-agnostic adversarial loss and optimization framework that aggregates adversarial and stealthiness loss gradients from multiple classifiers. Then, we propose an attention-based gradient weighting mechanism that concentrates perturbations on regions of high classification activation, thereby improving the robustness of adversarial projections when applied to scenes with varying camera poses. Our extensive experimental evaluations demonstrate that CAPAA achieves both a higher attack success rate and greater stealthiness compared to existing baselines. Codes are available at: \url{https://github.com/ZhanLiQxQ/CAPAA}.

\end{abstract}

\section{Introduction}
\label{sec:intro}

In multimedia security, adversarial attacks have emerged as a valuable approach to protect privacy and prevent the misuse of recognition systems. The development of such attacks has progressed from traditional methods like the Fast Gradient Sign Method (FGSM) \cite{Goodfellow2014ExplainingAH} to more sophisticated methods, such as attention-based \cite{huang2022attbased} and universal attacks \cite{2021universial}. 
Although these methods have made significant progress, they face challenges in real-world applications. As a result, researchers are increasingly exploring physical attacks---adversarial strategies that manipulate real-world objects or environments to deceive machine learning models, particularly in computer vision systems \cite{fang2024state}. An example is the attachment of special markers or stickers to objects \cite{Wei2021AdversarialSA}.

\begin{figure}[t]
\centering
\includegraphics[width=1.05\linewidth]{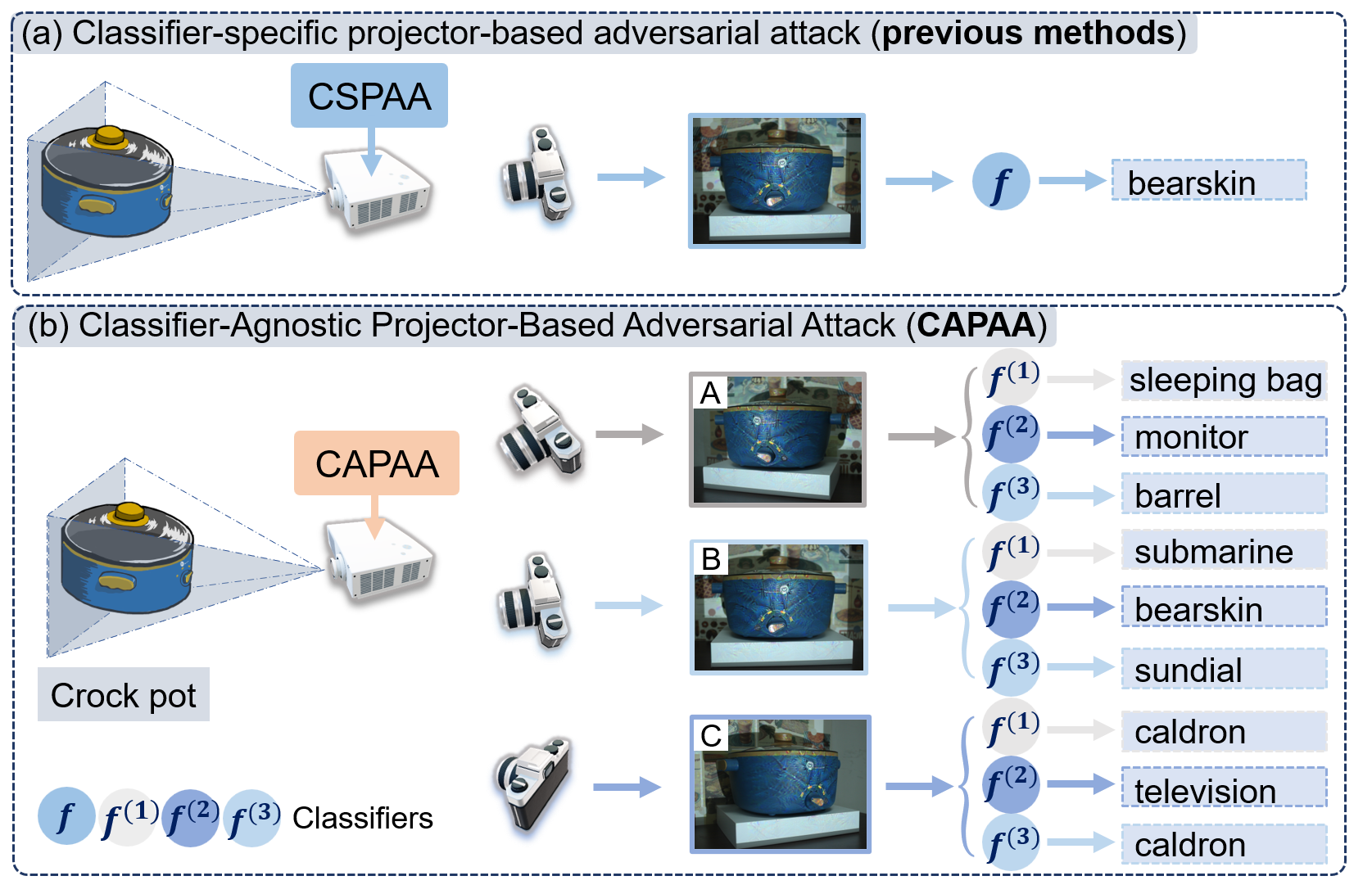}
\caption{\textbf{(a)} \textit{Classifier-specific} projector-based adversarial attack (CSPAA), aims to deceive a specific classifier under a specific camera capture pose by projecting adversarial light patterns. \textbf{(b)} \textit{Classifier-Agnostic} Projector-Based Adversarial Attack (CAPAA)  fools multiple classifiers simultaneously and is robust to camera pose changes. A real \textbf{Crock pot} (one of the ImageNet \cite{Deng2009ImageNetAL} classes) was placed in the scene, after projecting our CAPAA-generated adversarial light pattern, the camera-captured scene was misclassified by the three classifiers, such that their output labels were not \textbf{Crock pot}.} 
  \label{fig:teaser}
  \vspace{-12pt}
\end{figure}

\textbf{Projector-based attacks} are a form of physical attacks that deceive classifiers by manipulating illumination conditions without direct physical contact, as illustrated in Fig.~\ref{fig:teaser}~\textbf{(a)}. For instance, OPAD \cite{Gnanasambandam2021OpticalAA} exploits the optical interactions between projectors and cameras to execute attacks in real-world settings.

A key challenge for such attacks is achieving sufficient stealthiness, which is critical for their practical effectiveness. While methods like adversarial color projection \cite{hu2023adversarial} have been proposed, many struggle with this aspect. Recent metrics like hiPAA \cite{Wei2024PhysicalAA} provide a comprehensive evaluation framework by considering multiple factors, including effectiveness, robustness, and stealthiness.

While SPAA \cite{Huang2022SPAASP} improves stealthiness and robustness by modeling the project-and-capture process with a neural network, it remains limited to single-classifier scenarios with fixed camera poses. This restriction is particularly problematic given the growing use of ensemble classifiers \cite{guo2020ensemble}, as projector-based attacks optimized for a single classifier often fail to transfer effectively. Moreover, even minor camera pose perturbations can significantly degrade attack performance, a vulnerability that current pose-specific methods cannot easily overcome.

To overcome these challenges, we propose CAPAA (Classifier-Agnostic Projector-Based Adversarial Attack), a method designed to enhance attack robustness across various classifiers and camera poses. Specifically, for classifier-agnostic scenarios, CAPAA introduces a novel multi-objective loss function that enables joint attacks across multiple classifiers. Additionally, we incorporate attention-driven gradient weighting, which focuses subtle light perturbations on regions with high classification activation. These non-trivial designs improve the robustness and stealthiness of the attack.

Our contributions are summarized as follows:
\begin{itemize}
    \item To our best knowledge, CAPAA is the first classifier-agnostic, projector-based adversarial attack approach. 
    \item We introduce a new classifier-agnostic adversarial loss and optimization framework that aggregates adversarial and stealthiness loss gradients from multiple classifiers, allowing for more effective and flexible projector-based attacks across different classifiers.
    \item We propose an attention-based gradient weighting mechanism that focuses perturbations on regions of high classification activation, enhancing the robustness of adversarial projections even when camera pose changes.
    \item  Experimental evaluation across 10 setups and 7 camera poses demonstrates that CAPAA outperforms existing methods in terms of both stealthiness and success rates.
\end{itemize}
\section{Method}
\subsection{Problem formulation}

\noindent\textbf{Adversarial attacks}. Let $ f $ be an image classifier that maps an image $ I $ to a vector of $N$-class probabilities, $ f(I) \in [0, 1]^N $, where $ f_i(I) $ is the probability of the $ i $-th class. The goal of adversarial attack is to perturb the input image with almost imperceptible noise $\delta$, such that the classifier predicted class $\hat{y}$ either matches a target label $ y_t$ (targeted attack) or differs from the true label $y$ (untargeted attack):
{\small
\begin{align}\label{eq:adv_attack_formulation}
	&    \hat{y}= \underset{i}{\argmax} \, f_i(I+\delta) 
		\begin{cases}
		 = y_t, &\text{targeted}\\
		 \neq y,  & \text{untargeted}
		\end{cases}    \nonumber\\
        & \quad\quad\text{subject to} \quad \mathcal{D}\left(I, I+\delta\right) < \epsilon.
\end{align}
}\noindent The function $\mathcal{D}$ measures image similarity,  and is usually used to control the stealthiness of adversarial attack with a small threshold $\epsilon (\epsilon>0)$. 

\noindent\textbf{Projector-based adversarial attacks}. Extending \autoref{eq:adv_attack_formulation}
to the physical world that uses a projector to alter the light condition, and denote the physical scene as $s$, denote the projector's projection process, and the camera's capture process as $ \pi_p $ and $ \pi_c $, respectively. Then, given an input image $x$, the projected light of the projector can be expressed as $ \pi_p(x)$. In a specific camera pose $\gamma$, the scene captured by the camera under projected light can be represented as:
$I_{x,\gamma} = \pi_c(\pi_p(x), s, \gamma) $. For simplicity, we define the composite project-and-capture process as: 
$\pi(.) = \pi_c(\pi_p(.), s, \gamma)$, and we have $I_{x, \gamma} = \pi(x, \gamma)$.

Projector-based adversarial attacks aim to generate an adversarial image$/$pattern $x'$ as projector input, such that when projected to the physical scene and captured as $ I_{x', \gamma} $, it causes classifier $f$ to misclassify the scene: {\small
\begin{align}\label{eq:problem_formulation}
	&     \hat{y}= \underset{i}{\argmax} \, f_i(I_{x', \gamma}) 
		\begin{cases}
		 = y_t,  &\text{targeted}\\
		 \neq y,  & \text{untargeted}
		\end{cases}    \nonumber\\
        & \quad\quad\text{subject to} \quad \mathcal{D}\left(I_{x', \gamma}, I_{x_0, \gamma}\right) < \epsilon,
\end{align}
}where $I_{x_0, \gamma}$ is the camera-captured scene illuminated by gray light $x_0$, i.e., without adversarial projection. Previous \textbf{classifier-specific} methods \cite{Huang2022SPAASP, Gnanasambandam2021OpticalAA} are based on the formulation in \autoref{eq:problem_formulation}. Although straightforward, they may fail when applied to other classifiers, because the adversarial projection is generated using feedback from a specific classifier. Furthermore, as adversarial projections may become occluded, they may also fail when the camera pose $\gamma$ changes. 

\subsection{Classifier-Agnostic Projector-Based Adversarial Attack \\(CAPAA)}
\begin{figure*}[ht]
\centering
\includegraphics[width=\linewidth]{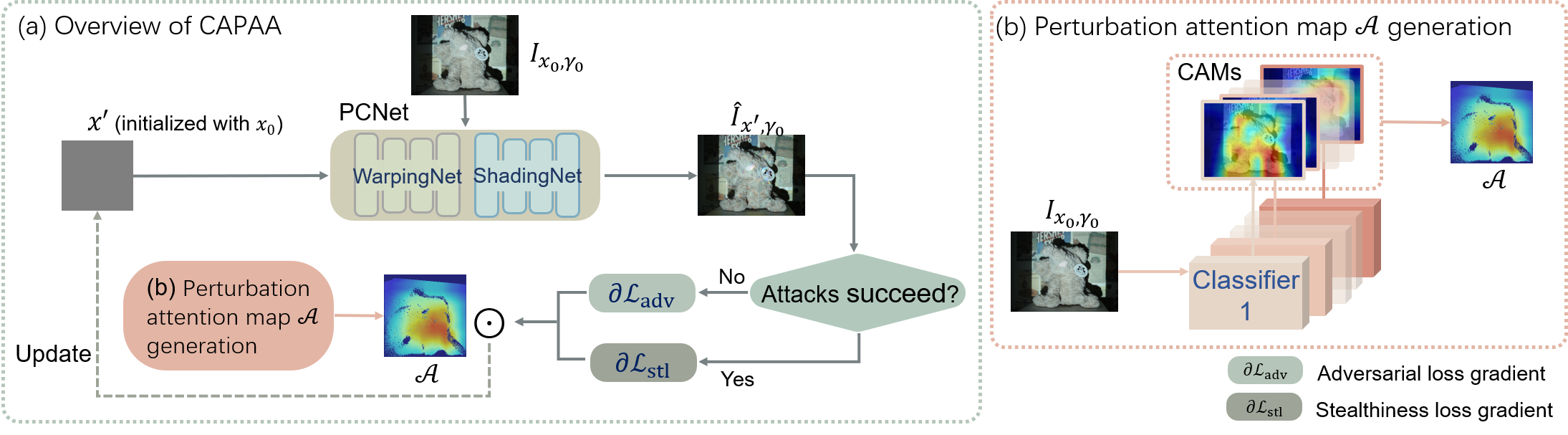}
\caption{\textbf{(a)} Overview of CAPAA. We first input the adversarial projection $x'$ (initialized with gray image $x_0$) and the camera image $I_{x_0,{\gamma}_0}$ to the trained PCNet to obtain the inferred projection $\hat I_{x',\gamma_0}$. After generating perturbation attention maps (PAM) for each classifier, we calculate their weighted sum $\mathcal{A}$ for attention-based gradient weighting. The optimization follows an alternating mechanism, i.e., if $\hat{I}_{x',{\gamma}_0}$ successfully attacks the classifiers, the stealthiness loss gradient is calculated and weighted by $\mathcal{A}$ to update $x'$. Otherwise, the classifier-agnostic adversarial loss gradient is applied to update $x'$, as outlined in Algorithm 1. 
\textbf{(b)} Perturbation attention map $\mathcal{A}$ generation. For each classifier, we first generate its class activation map (CAM) of the camera-captured scene image $I_{x_0, \gamma_0}$ using Grad-CAM++ \cite{Chattopadhyay2017GradCAMGG}. The weighted sum of these individual CAMs is utilized as our PAMs' $\mathcal{A}$, enabling our CAPAA to generate adversarial projections towards the most salient regions of the object.
}
\vspace{-14pt}
\label{fig:network}

\end{figure*}

To address the issues above, we propose CAPAA to generate adversarial projection $x'$ that can perform \textit{classifier-agnostic} attack, and still be robust when camera pose changes:
{\small
\begin{align}\label{eq:pappa_formulation}
	  &\forall f^{(k)} \in \{f^{(1)}, f^{(2)}, ..., f^{(n)}\} \nonumber\\
       &\hat{y}^{(k)}= \underset{i}{\argmax} \, f^{(k)}_i(I_{x', \gamma}) 
		\begin{cases}
		 = y_t,  &\text{targeted}\\
		 \neq y,  & \text{untargeted}
		\end{cases}    \nonumber\\
        & \quad\quad\text{subject to} \quad \mathcal{D}\left(I_{x', \gamma}, I_{x_0, \gamma}\right) < \epsilon,
\end{align}
}\noindent where {\small$f^{(k)} \in \{f^{(1)}, f^{(2)}, ..., f^{(n)}\}$} is the $k$-th classifier to be attacked. 
To ensure robust and stealthy attacks, we alternatively minimize adversarial and stealthiness losses below:
\begin{equation}\label{eq:attack_loss} 	
{\small  x' = \argmin_{x'} \alpha \mathcal{L}_{\rm adv}\left(\hat{I}_{x', \gamma}\right) +  		\mathcal{D}\left(\hat{I}_{x', \gamma}, ~I_{x_0, \gamma}\right), }
\end{equation}
where $\alpha=-1$ for targeted attacks and $\alpha=1$ for untargeted attacks. 
$\mathcal{D}$ is perceptual color distance $\Delta E$ (\ie, CIEDE2000 \cite{luo2001development}), and it has been experimentally demonstrated to better align with human visual perception and produce more robust and transferable attacks \cite{zhao2020towards} compared with $l_p$ norm. $\hat{I}_{x', \gamma}$ represents the simulated camera-captured adversarial projection rather than the real one ($I_{x', \gamma}$) to avoid including the physical project-and-capture process $\pi$ in the optimization loop because $\pi$ is non-differentiable and it is highly inefficient even with gradient-free optimization. Inspired by \cite{Huang2022SPAASP}, we use a neural network named PCNet $\hat{\pi}_\theta$ (parameterized by $\theta$) to approximate the physical project-and-capture process $\pi$. 
PCNet consists of two components: ShadingNet (for photometry) and WarpingNet (for geometry), as shown in \autoref{fig:network}. The simulated project-and-capture process is denoted as $\hat{I}_{x', \gamma} = \hat{\pi}_{\theta, \gamma}(x')$ , with $\theta$ representing its parameters. PCNet is trained by minimizing the loss between the real captured projections $I_{x,\gamma}$ and the inferred ones $\hat{I}_{x,\gamma}$:
\begin{equation}{\small\label{eq:pcnet_loss} 	\theta= \argmin_{\theta}\sum\nolimits_i\mathcal{L}_{\rm PC}\big(\hat{I}_{x_i,\gamma_0} = \hat{\pi}_{\theta', \gamma_0}(x_i), ~I_{x_i,\gamma_0}\big) ,}\end{equation} 
where {\small$\mathcal{L}_{\rm PC}$} is pixel-wise {\small$L_1+\text{DSSIM}$} loss, $\gamma_0$ is the camera pose where PCNet is trained, and {\small$\{(x_i, I_{x_i, \gamma_0})\}_{i=1}^{M}$ } forms $M$ pairs of real projected and captured images for training.

\noindent\textbf{Classifier-Agnostic adversarial loss.}\label{subsec:multi-class-loss}
We now introduce the adversarial loss function $\mathcal{L}_{\rm adv}$ for classifier-agnostic attacks. For classifier-agnostic \textbf{untargeted} attacks, an intuitive solution is to use the weighted sum of the adversarial loss of each classifier. Denote {\small $z^{(k)}_i(\cdot)$} as the $k$-th classifier's output logit (raw classification score) of the $i$-th label, which is related to $f^k_i$ by: {\small $f^k_i = \text{softmax}(z^{(k)}_i)$}. Then, our untargeted classifier-agnostic adversarial attack loss is given by
{\small
\begin{align}\label{eq:untar_multi_class_loss}
 	&   \mathcal{L}_{\rm adv}\left(\hat{I}_{x', \gamma}\right) = \sum\nolimits_{k = 1}^{n} \omega_{k} \cdot z^{(k)}_i\left(\hat{I}_{x', \gamma_0}\right),
\end{align}
}where $\omega_{k}$ stands for the weight of the $k$-th classifier.

A more challenging problem is the \textbf{targeted} attack, where the above simple weighted sum of the adversarial loss of each classifier may fail, as the simulated projector-based attack may fail in the real world, due to the perturbations of the complex environment. In such cases, the classifier may recognize the real camera-captured object under the adversarial projection as neither the object's true class nor the attack's target class, but rather as a class similar to the target class. For example, when projecting an adversarial pattern onto the object \textbf{Teddy} to fool the classifier into recognizing it as \textbf{rooster}, the classifier might instead output \textbf{hen}. This is because the original softmax function inherently emphasizes the largest logit, and adversarial attacks may produce right-above-the-margin perturbations, which are less robust after real-world project-and-capture processes. 
To address this issue, we add stricter constraints to the classifier's output logits by controlling the temperature of the LogSoftmax function, such that the adversarial attack is only successful when the classifier's target logit is significantly higher than the other classes:
\vspace{-0.3em}
{\small\begin{equation}\label{eq:tar_multi_class_loss}
{\mathcal{L}_{\rm adv}\left(\hat{I}_{x', \gamma}\right) = \sum\nolimits_{k = 1}^{n} \omega_{k} \cdot \text{LogSoftmax} \left(z^{(k)}_{t}\left(\hat{I}_{x', \gamma_0}\right) / T\right),
}\end{equation}}where the parameter $T$ acts as a temperature parameter, and it is dynamically adjusted during the optimization process. When  $T = 1 $, the softmax function behaves as a standard output layer for classifiers. For $ T > 1 $, the Softmax distribution becomes smoother. In adversarial training, this helps classifiers with standard Softmax outputs generate adversarial examples that better distinguish between the target class and similar classes (e.g., hen and rooster), thereby reducing ambiguity.

\begin{figure*}[tb]
  \centering
  \includegraphics[width=1\linewidth]{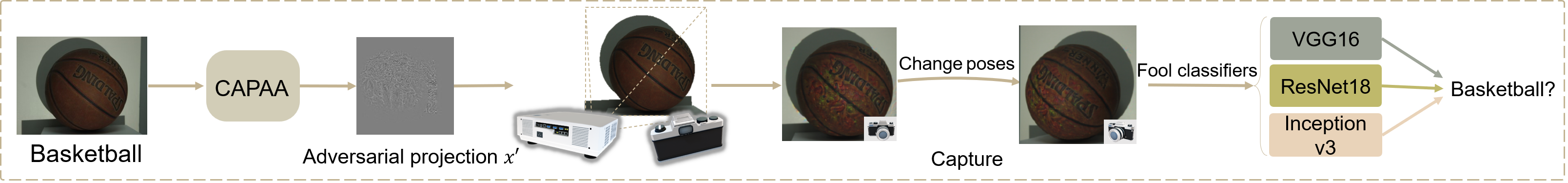}
  \caption{Overview of the experimental evaluation. First, we sample the object and train PCNet. Then, we use different methods (\eg, CAPAA) to generate the adversarial projections. After that, we project the adversarial patterns onto the object and move the camera to capture the scene in different poses. Finally, the captured images (the object with superimposed adversarial projection) are fed to different classifiers for prediction.}
  \label{fig:experiment_procedure}
  \vspace{-12pt} %
  \end{figure*}

\noindent\textbf{Attention-based gradient weighting.}\label{subsec:attention-gradient} 
To improve the robustness of the adversarial projection under varying camera poses, we propose an attention-based gradient weighting mechanism. It is based on the observation that (1) adversarial projections may be occluded or move out of the camera's field of view when the camera pose changes. However, most existing methods apply perturbations uniformly across all regions, and may fail when the camera pose changes. (2) Classifiers often focus on specific regions of the object when making predictions. Therefore, we propose to focus perturbations on regions with strong classification activation, as shown in Fig.~\ref{fig:network}~\textbf{(b)}.

To find the regions of strong classification activation, an intuitive method is to use an object detector, such as YOLO \cite{Redmon2016YouOL}, to locate the object and apply perturbations within its bounding box. However, this introduces additional complexity and potential reliability issues with detection. Instead, we employ an attention mechanism, specifically Grad-CAM++ \cite{Chattopadhyay2017GradCAMGG}, to find the class activation map (CAM) on the object. Then, in each adversarial attack iteration, we weigh the loss gradient using CAM, focusing the perturbations on regions with high classification activation, as shown in \autoref{eq:CAM_CAPAA_loss}.
{\footnotesize
\begin{equation}\label{eq:CAM_CAPAA_loss}
\begin{split}
\frac{\partial\mathcal{L}_\text{CAPAA}}{\partial x'} = \mathcal{A} \odot \Bigg(
\underbrace{\frac{\partial\mathcal{L}_{\rm adv}\left(\hat{I}_{x', \gamma_0}\right)}{\partial x'}}_{\text{adversarial loss gradient}} + \underbrace{\frac{\partial\mathcal{L}_{\rm stl}\left(\hat{I}_{x', \gamma_0}, ~I_{x_0, \gamma_0}\right)}{\partial x'}}_{\text{stealthiness loss gradient}
 }\Bigg), \end{split}
\end{equation}
}where $\mathcal{A}$ is the perturbation attention map (PAM) represented by CAM, and $\odot$ denotes element-wise multiplication. 
The overall process of CAPAA is illustrated in Fig.~\ref{fig:network} and Algorithm~\ref{alg:algorithm1}. To elucidate, we first initialize $ x' $ with a plain gray projector image $ x_0 $ and set $\mu=1/N$ for each classifier's PAM $\mathcal{A}^{(k)}$. We set the learning rate $\beta_1 = 2$ for minimizing the adversarial loss and $\beta_2 = 1$ for minimizing the stealthiness loss. We then iteratively update $ x' $ by minimizing the adversarial loss when the adversarial confidence is below a threshold $ p_{\rm thr} =0.9$ or the perturbation size is below a threshold $ d_{\rm thr}\ (2 \leq d_{\rm thr} \leq 5$). Otherwise, we minimize the stealthiness loss. The final output adversarial projection $ x' $ is the one that is adversarial and has the smallest perceptual color distance $\Delta E$ to the original projector image $ x_0 $.
\setlength{\algomargin}{0em}
\begin{algorithm}[t]
\small 
  \SetAlCapHSkip{0em}
  \SetAlgoLined 
  \SetKwComment{Comment}{$\triangleright$\ }{} 
  \SetSideCommentLeft
  \SetAlgoLongEnd
  \DontPrintSemicolon
  \SetKwInput{Input}{Input}
  \SetKwInOut{Output}{Output}
  \SetKw{KwOr}{or}
  \caption{CAPAA: Classifier-Agnostic Projector-Based Adversarial Attack}\label{alg:algorithm1}
  \Input{\;
  
$x_0, I_\text{m}$: projector plain gray image, projector direct light mask\;
$I_s$: camera-captured scene under $x_0$ projection\;
$\mathcal{A}$: perturbation attention maps (PAM)\; 
$\mu^{(k)}$: weight of the $k$-th classifier's PAM\;
$K$: number of iterations\;
$p_{\rm thr}$: threshold for adversarial confidence\;
$d_{\rm thr}$: threshold for $\Delta E$ perturbation size\;
$\beta_1,\beta_2$: step sizes for adversarial and stealthiness losses\;
  }
  \Output{$x'$: projector input adversarial image}
  
  
  \vspace{1.5mm}Initialize $x'_0 \gets x_0$\;

  $\mathcal{A} \gets \sum_{k = 1}^{N}\mu^{(k)} \mathcal{A}^{(k)}$\;
  
  \For{$j\gets1$ \KwTo $K$}
  {
    $\hat{I}_{x',\gamma_0} \gets \hat{\pi}_{\theta, \gamma_0}(x'_{j-1})$\;

    $d \gets \mathcal{D}\left(\hat{I}_{x',\gamma_0}, I_{x_0, \gamma_0}\right)$\;

  \eIf{$f_{y_t}(\hat{I}_{x',\gamma_0}) < p_{\rm thr}$ \KwOr $d<d_{\rm thr}$}     
      {
      $g_1\gets \mathcal{A} \odot \alpha\nabla_{x'} \mathcal{L}_{\rm adv}\left(\hat{I}_{x', \gamma_0}\right)$ {\footnotesize \hfill // min. adversarial loss\;}
      $x'_j\gets x'_{j-1}+ \beta_1*\frac{g_1}{\norm{g_1}_2}$\;
      }
      {
       $g_2 \gets - \mathcal{A} \odot \nabla_{x'} d $  {\footnotesize \hspace{6em}\hfill// min. stealthiness loss\;}
      $x'_j \gets x'_{j-1}+\beta_2*\frac{g_2}{\norm{g_2}_2}$\;
      }
      $x'_j \gets \text{clip}(x'_j, 0, 1)$\; 
  }
  \KwRet $x' \gets x'_j$ that is adversarial and has smallest $d$\;
\end{algorithm}
\section{Experimental Evaluation}\label{sec:experiments}
\subsection{Experiment setup}
As shown in Fig.~\ref{fig:experiment_procedure}, our setup consists of a projector and a camera, both facing a target object to be attacked. We start by capturing the object image under gray light $x_0$ and training PCNet. We then generate adversarial patterns using four different methods, including CAPAA, for both targeted (10 targets) and untargeted attacks. Next, we project the generated adversarial patterns onto the object and capture the scene under different camera poses, e.g., the original pose, different angles (±15°, ±30°) and different focal lengths (±5mm). Finally, we feed the camera-captured images into three classifiers (ResNet-18 \cite{He2016DeepRL}, VGG-16 \cite{Simonyan2015VeryDC}, and Inception v3 \cite{Szegedy2016RethinkingTI}) for real-world projector-based adversarial attack evaluation.

\noindent\textbf {Evaluation metrics.} 
To measure the attack success rate and stealthiness, we define a stealthiness-constrained attack success rate metric for the camera-capture adversarial projection $I_{x', \gamma}$:
\vspace{-1.5em}
{\small
\begin{align}
    \mathcal{S}^{(k)}_h(I_{x', \gamma}) =  \begin{cases}
    1, \quad {\rm if~~}\hat{y}= \underset{i}{\argmax} \, f_i(I_{x', \gamma})  \begin{cases}
		 = y_t, \quad \quad \text{targeted}\\
		  \neq y, \quad \text{untargeted}
		\end{cases}  \\ \quad\quad\quad\quad
    \text{and~}  \mathcal{D}\left(I_{x', \gamma},I_{x_0, \gamma}\right)\leq h
        \quad  \quad \\ 
     0, \quad \text{otherwise}. \end{cases}
\nonumber
\end{align}
}This metric ensures that a projector-based attack is successful only when it fools the given classifier and its stealthiness $\Delta E$ is no greater than $h$. Then, we plot the success rate vs stealthiness diagrams of all compared methods. As shown in \autoref{fig:all_quantity}\textbf{(a) - (c)}, the horizontal axis corresponds to the perturbation threshold $\Delta E$\cite{luo2001development}, and the vertical axis represents the cumulative success rate $\mathcal{C}_h$ at a given $\Delta E$ threshold $h$:
{\small
\begin{align}\label{eq:cumulative_success_rate}
	& \mathcal{C}_h = \frac{1}{PNH}\sum\nolimits_{j = 0}^{P-1}\sum\nolimits_{k = 1}^{N}\sum\nolimits_{l = 1}^{H}\mathcal{S}_h^{(k)}(I_{x'_l, \gamma_j}),
\end{align}
}where $P$, $N$, $H$ are the number of camera poses, the number of image classifiers to be attacked, and the number of generated adversarial perturbations, respectively. In particular, {\small $\mathcal{S}_h^{(k)}(I_{x'_l, \gamma_j})$} indicates whether the $l$-th camera-captured adversarial projection successfully fools the $k$-th classifier $f^{(k)}$ at the $j$-th camera pose, meanwhile, its $\Delta E$ is less than $h$. 
Note that we evaluate: (i) targeted attacks at the original pose ($P = 1$, \autoref{fig:all_quantity}~\textbf{(b)}), 
and (ii) targeted/untargeted attacks across multiple poses ($P = 7$, \autoref{fig:all_quantity}~\textbf{(a)} \& \textbf{(c)}).

\begin{table}[!t]
\vspace{-10pt}
\begin{flushright}
\caption{ 
Quantitative comparisons for classifier-agnostic multi-pose \textbf{untargeted} attacks. Four stealthiness thresholds $d_{\rm thr} \in \{2,3,4,5\}$ are used to generate adversarial projections (2nd column). Columns 3 to 6 present stealthiness metrics for \textit{camera-captured} adversarial projections, column 7 indicates the average top-1 success rate, and column 8 shows the average top-1 success rate \textit{across all stealthiness thresholds} over 10 different setups, and each setup consists of 7 camera poses.}
\label{tab:table 1}
\vspace{-10pt}
\begin{tabular}{@{}l@{}c@{\hspace{7pt}}c@{\hspace{7pt}}c@{\hspace{7pt}}c@{\hspace{7pt}}c@{\hspace{7pt}}c@{}c@{}} 
	\toprule[0.5mm]
\textbf{Attacker}                                                                                                 & \textbf{d$_\textbf{thr}$} & \textbf{L$_\textbf{inf}\downarrow$} & \textbf{L$_\textbf{2}\downarrow$} & \textbf{$\Delta \textbf{E}\downarrow$} & \textbf{SSIM$\uparrow$} & \textbf{U.top-1} & \textbf{\begin{tabular}[c]{c@{}}Avg. attack \\success rate\end{tabular}}     \\ \midrule
\multirow{4}{*}{\begin{tabular}[c]{@{}l@{}}{SPAA} \cite{Huang2022SPAASP} \end{tabular} }                             & 2 & 5.11                                     & 6.38                                   & 2.25                                   & 0.914                                     & 51.43\%                                    & \multirow{4}{*}{64.68\%}                                                                           \\
                                                   & 3 & 7.16                                     & 8.94                                   & 3.01                                   & 0.862                                     & 62.86\%                                    &                                                                                                    \\
                                                   & 4 & 9.02                                     & 11.18                                  & 3.83                                   & 0.828                                     & 68.73\%                                    &                                                                                                    \\
                                                   & 5 & 10.64                                    & 13.08                                  & 4.63                                   & 0.805                                     & 75.71\%                                    &                                                                                                    \\ \midrule
\multirow{4}{*}{\begin{tabular}[c]{@{}l@{}} CAPAA w/o \\attention\end{tabular}}               & 2                     & 5.24                                     & 6.55                                   & 2.29                                   & 0.911                                     & 71.90\%                                    & \multirow{4}{*}{\textbf{82.02\%}}                                                                  \\
                                                   & 3                     & 7.23                                     & 9.04                                   & 3.05                                   & 0.860                                     & 81.43\%                                    &                                                                                                    \\
                                                   & 4                     & 9.04                                     & 11.20                                  & 3.87                                   & 0.827                                     & 87.14\%                           &                                                                                                    \\
                                                   & 5                     & 10.56                                    & 12.98                                  & 4.66                                   & 0.804                                     & \textbf{87.62\%}                           &                                                                                                    \\ \midrule
\multirow{4}{*}{\begin{tabular}[c]{@{}l@{}} CAPAA \\classifier-specific \\\end{tabular}} & 2                     & \textbf{4.73}                            & \textbf{5.89}                          & \textbf{2.09}                          & 0.927                                     & 51.59\%                                    & \multirow{4}{*}{61.75\%}                                                                           \\
                                                   & 3                     & 6.43                                     & 7.96                                   & 2.80                          & 0.889                                     & 62.70\%                                    &                                                                                                    \\
                                                   & 4                     & 7.72                                     & 9.47                                   & 3.47                                   & 0.868                                     & 65.08\%                                    &                                                                                                    \\
                                                   & 5                     & 8.49                                     & 10.37                                  & 3.92                                   & 0.858                                     & 67.94\%                                    &                                                                                                    \\ \midrule
\multirow{4}{*}{CAPAA (ours)}                      & 2                     & 4.77                                     & 5.95                                   & 2.10                                   & \textbf{0.930}                            & 74.76\%                           & \multirow{4}{*}{\textbf{82.02\%}}                                                                           \\
                                                   & 3                     & 6.36                            & 7.85                          & 2.82                                   & 0.895                            & 81.90\%                           &                                                                                                    \\
                                                   & 4                     & 7.48                            & 9.15                          & 3.46                          & 0.877                            & 84.76\%                                    &                                                                                                    \\
                                                   & 5                     & 8.01                            & 9.74                          & 3.83                          & 0.871                            & 86.67\%                                    &                                                                                                    \\ 
                                                   \bottomrule[0.5mm]
\end{tabular}
\end{flushright}
  \hfill 
  \vspace{-20pt} 
\end{table}

\begin{figure*}[ht]
\centering
\includegraphics[width=1\linewidth]{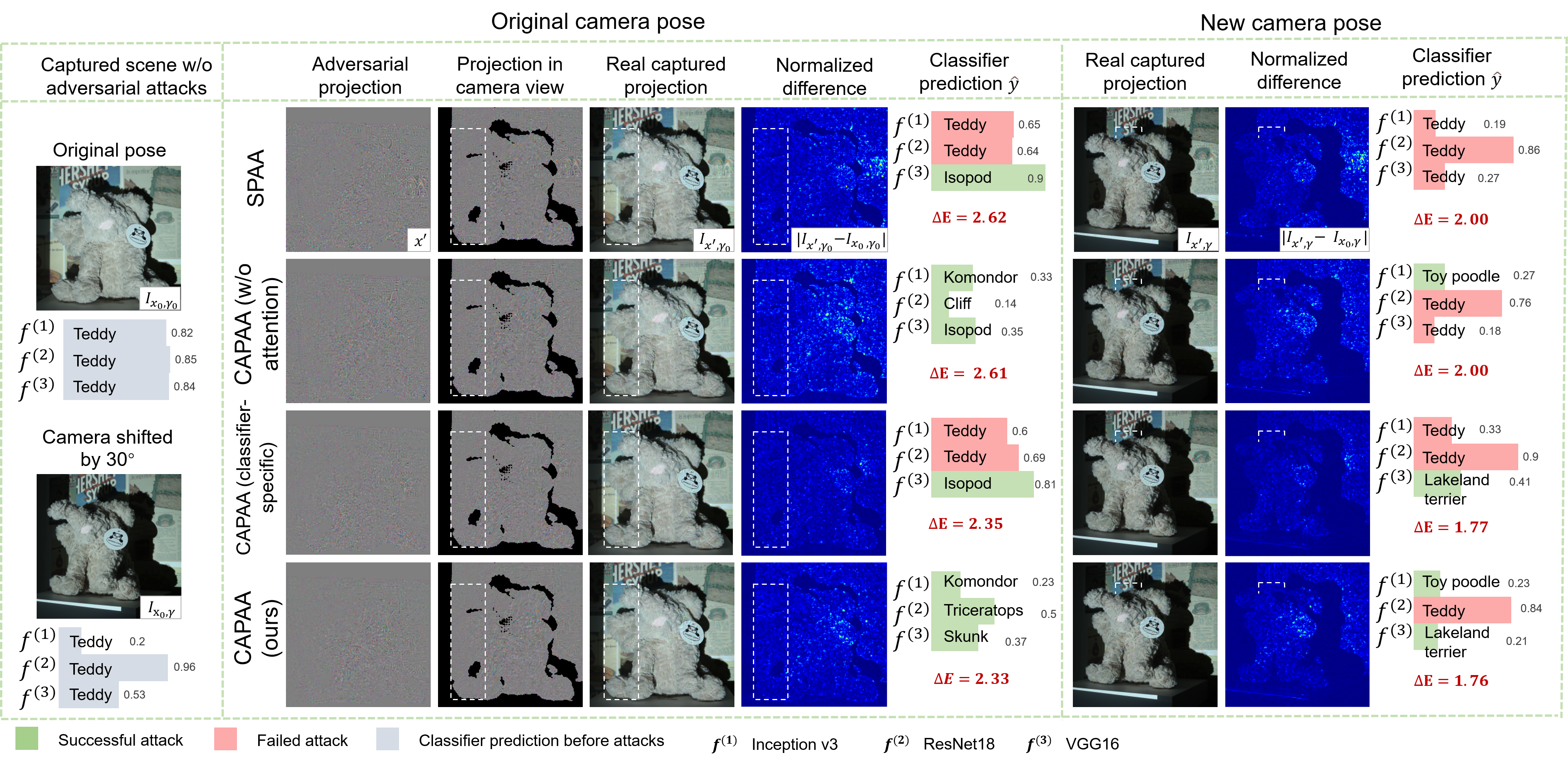}
\caption{Qualitative comparisons of classifier-agnostic \textbf{untargeted} attacks across two camera views. The classifier prediction $\hat{y}$, including the probabilities, is displayed on the bottom or right side of each image. The perturbations highlighted by the white dashed boxes, especially in the 5th and 7th columns, indicate that attention-based attacks, CAPAA and CAPAA (classifier-specific) 
tend to avoid attacking background regions due to the CAM mechanism, and thus are more robust against occlusions (caused by camera pose changes) compared to other baselines.
}
\label{fig:quality_teddy}
\vspace{-14pt} 
\end{figure*} 
\noindent\textbf{Compared baselines.} 
We compare our CAPAA with three baselines: SPAA~\cite{Huang2022SPAASP}, CAPAA (w/o attention), and CAPAA (classifier-specific). SPAA~\cite{Huang2022SPAASP} is the closest projector-based adversarial attack method to our CAPAA, but it is classifier-specific and does not consider attack robustness across other camera poses. CAPAA (w/o attention) is a degraded CAPAA that jointly attacks multiple classifiers but with no attention-based gradient weighting, and CAPAA (classifier-specific) is a degraded CAPAA without classifier-agnostic adversarial loss, thus can only attack each classifier individually. Since SPAA and CAPAA (classifier-specific) cannot perform classifier-agnostic attacks, we evaluate them in a classifier-specific manner, i.e., attack each classifier individually, which gives them advantages over classifier-agnostic ones.

Clearly, CAPAA outperforms all baseline approaches in terms of stealthiness.

\vspace{-3pt}
\subsection{Experimental results}
\begin{figure}[h]
\centering
\includegraphics[width=1\linewidth]{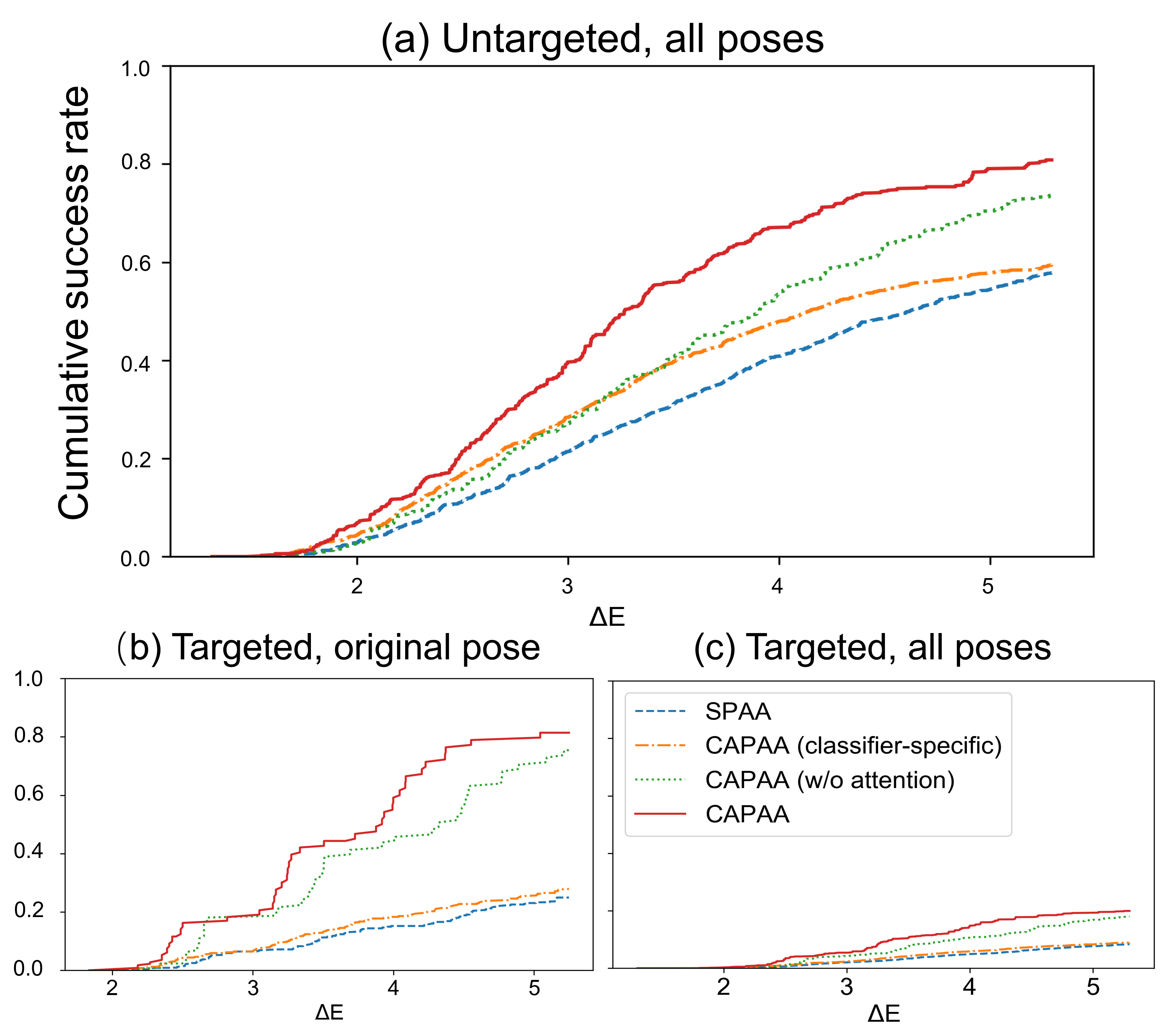}
\caption{Quantitative comparisons on projector-based classifier-agnostic adversarial attacks. \textbf{(a)}  \textbf{Untargeted} attacks. \textbf{(b)} \textbf{Targeted} attacks under the \textit{original} camera capture pose. \textbf{(c)} \textbf{Targeted} attacks across\textit{ all} camera capture poses. } 
\label{fig:all_quantity}
\vspace{-14pt} 
\end{figure}

\noindent\textbf{Untargeted attack.} 
As shown in \autoref{tab:table 1}, the average attack success rates of CAPAA and CAPAA (w/o attention) achieve the highest attack success rates (with a marginal 0.001\% difference) and consistently outperform other methods across various stealthiness thresholds. Moreover, CAPAA outperforms CAPAA (w/o attention) when the stealthiness threshold $d_\text{thr} \leq 3$ and excels in stealthiness metrics such as $L_\text{inf}$, $L_{2}$, $\Delta E$\cite{luo2001development}, and SSIM. CAPAA (classifier-specific) enhances stealthiness. Notably, CAPAA maintains high success rates while enhancing stealthiness, demonstrating its capability to generate more robust adversarial projections. The curves in \autoref{fig:all_quantity}~\textbf{(a)} further indicate that CAPAA shows the most rapid growth and the highest cumulative success rate, underscoring its effectiveness in balancing stealthiness and success rate.
\begin{figure*}[!t]
\centering
\includegraphics[width=1.04\linewidth]{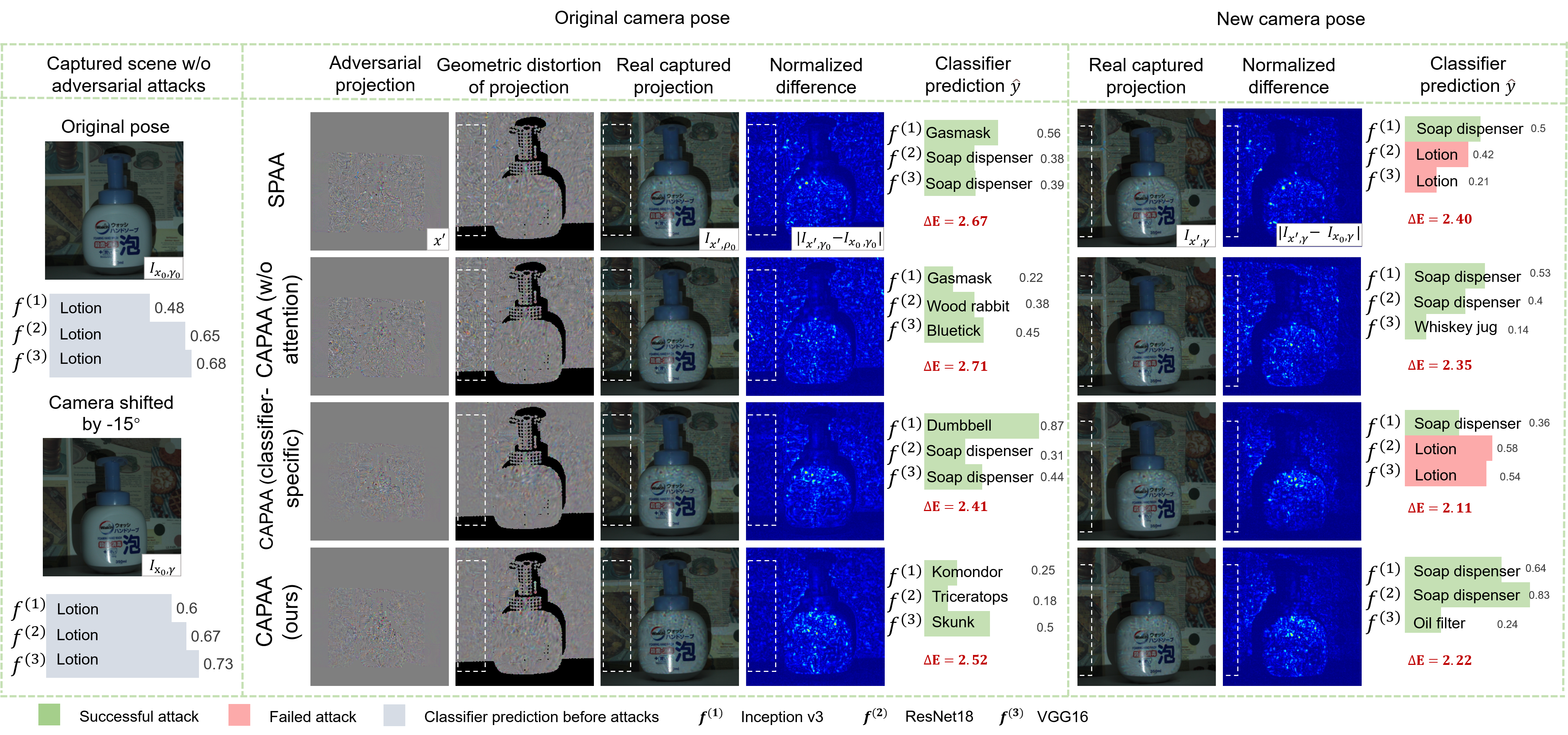}
\caption{Qualitative results of the classifier-agnostic and multi-pose \textbf{untargeted} attacks. Specifically, the white frames show how the perturbations on the background are out of the camera FOV after shifting the camera angle.}
\label{fig:quality_untargeted}
\vspace{-15pt} 
\end{figure*} 

Note that after changing camera poses, some adversarial projections become invisible due to occlusion. For example, in \autoref{fig:quality_untargeted}, the background perturbations highlighted in white dashed boxes are out of the camera FOV after changing the camera pose. In \autoref{fig:quality_teddy}, the background perturbations are occluded by the object \textbf{Teddy} after changing the camera pose. However, CAPAA and CAPAA (classifier-specific) are less affected because they can focus adversarial perturbations on the object by using CAM.  Moreover, attention-based techniques yield stealthier projections; for example, CAPAA (classifier-specific) and CAPAA exhibit smaller stealthiness ($\Delta E$) than other baselines in the original pose. 
Notably, in Fig.~\ref{fig:quality_teddy}, after a 30° camera shift, only CAPAA achieved two successful attacks with the highest stealthiness, while SPAA failed in all attempts.
Similarly, as shown in the attacks against \textbf{Lotion} in Fig.~\ref{fig:quality_untargeted}, SPAA only succeeded once, whereas CAPAA successfully fooled all classifiers with a smaller $\Delta E$. Although CAPAA (w/o attention) also succeeded, but with a higher $\Delta E$. We also conducted additional experiments attacking Vision Transformers (ViTs) \cite{dosovitskiy2020} and four unseen classifier architectures. The results demonstrate consistent superiority over baselines across all tested models, while revealing limitations for future improvement (details are in the supplementary material).

\noindent\textbf{Targeted attack.} 
\autoref{fig:all_quantity}~\textbf{(c)} shows that CAPAA outperforms other methods in both success rate and stealthiness on targeted attacks. Note that targeted attacks are much more challenging than untargeted ones, resulting in lower average success rates. CAPAA and CAPAA (w/o attention) lead in performance for classifier-agnostic targeted attacks at the original camera pose, with CAPAA (w/o attention) tripling the success rate due to the three classifiers targeted. CAPAA also shows improved performance when lower stealthiness (i.e., larger $\Delta E$) is allowed, confirming its effectiveness, particularly at the original camera pose (\autoref{fig:all_quantity}~\textbf{(b)}).

\vspace{-1pt}
\section{Conclusion and limitations}\label{sec:conclusion}
We propose CAPAA, a classifier-agnostic projector-based adversarial attack method that is robust even when the camera pose changes. CAPAA combines a novel classifier-agnostic adversarial loss with an attention-based gradient weighting strategy to achieve both stealthy and robust adversarial projections. On a benchmark with 10 setups (10 objects and 7 poses), we show that CAPAA outperforms existing methods in stealthiness and achieves high attack success rates.

\noindent\textbf{Limitations and future work.}
Although robust against camera pose changes, CAPAA is not pose-agnostic because it does not aggregate attack loss gradients from multiple camera poses. Future work is to incorporate various camera poses to address this issue. 
\vspace{-0.5em}
\bibliographystyle{IEEEtran}
\bibliography{icme2025ref}

\begin{thebibliography}{10}
\providecommand{\url}[1]{#1}
\csname url@samestyle\endcsname
\providecommand{\newblock}{\relax}
\providecommand{\bibinfo}[2]{#2}
\providecommand{\BIBentrySTDinterwordspacing}{\spaceskip=0pt\relax}
\providecommand{\BIBentryALTinterwordstretchfactor}{4}
\providecommand{\BIBentryALTinterwordspacing}{\spaceskip=\fontdimen2\font plus
\BIBentryALTinterwordstretchfactor\fontdimen3\font minus \fontdimen4\font\relax}
\providecommand{\BIBforeignlanguage}[2]{{%
\expandafter\ifx\csname l@#1\endcsname\relax
\typeout{** WARNING: IEEEtran.bst: No hyphenation pattern has been}%
\typeout{** loaded for the language `#1'. Using the pattern for}%
\typeout{** the default language instead.}%
\else
\language=\csname l@#1\endcsname
\fi
#2}}
\providecommand{\BIBdecl}{\relax}
\BIBdecl

\bibitem{Goodfellow2014ExplainingAH}
I.~J. Goodfellow, J.~Shlens, and C.~Szegedy, ``Explaining and harnessing adversarial examples,'' \emph{ICLR}, vol. abs/1412.6572, 2015.

\bibitem{huang2022attbased}
Q.~Huang, Z.~Lian, and Q.~Li, ``Attention based adversarial attacks with low perturbations,'' in \emph{ICME}, 2022, pp. 1--6.

\bibitem{2021universial}
P.~Benz, C.~Zhang, A.~Karjauv, and I.~S. Kweon, ``Universal adversarial training with class-wise perturbations,'' in \emph{ICME}, 2021, pp. 1--6.

\bibitem{fang2024state}
J.~Fang, Y.~Jiang, C.~Jiang, Z.~L. Jiang, C.~Liu, and S.-M. Yiu, ``State-of-the-art optical-based physical adversarial attacks for deep learning computer vision systems,'' \emph{ESWA}, p. 123761, 2024.

\bibitem{Wei2021AdversarialSA}
X.~Wei, Y.~Guo, and J.~Yu, ``Adversarial sticker: A stealthy attack method in the physical world,'' \emph{TPAMI}, vol.~45, pp. 2711--2725, 2021.

\bibitem{Deng2009ImageNetAL}
J.~Deng, W.~Dong, R.~Socher, L.-J. Li, L.~Kai, and F.-F. Li, ``Imagenet: A large-scale hierarchical image database,'' in \emph{CVPR}, 2009, pp. 248--255.

\bibitem{Gnanasambandam2021OpticalAA}
A.~Gnanasambandam, A.~M. Sherman, and S.~H. Chan, ``Optical adversarial attack,'' \emph{ICCVW}, pp. 92--101, 2021.

\bibitem{hu2023adversarial}
C.~Hu, W.~Shi, and L.~Tian, ``Adversarial color projection: A projector-based physical-world attack to dnns,'' \emph{Image and Vision Computing}, vol. 140, p. 104861, 2023.

\bibitem{Wei2024PhysicalAA}
H.~Wei, H.~Tang, X.~Jia, Z.~Wang, H.~Yu, Z.~Li, S.~Satoh, L.~Van~Gool, and Z.~Wang, ``Physical adversarial attack meets computer vision: A decade survey,'' \emph{TPAMI}, vol.~46, no.~12, pp. 9797--9817, 2024.

\bibitem{Huang2022SPAASP}
B.~Huang and H.~Ling, ``Spaa: Stealthy projector-based adversarial attacks on deep image classifiers,'' in \emph{VR}, 2022, pp. 534--542.

\bibitem{guo2020ensemble}
Y.~Guo, X.~Wang, P.~Xiao, and X.~Xu, ``An ensemble learning framework for convolutional neural network based on multiple classifiers,'' \emph{Soft Computing}, vol.~24, no.~5, pp. 3727--3735, 2020.

\bibitem{luo2001development}
M.~R. Luo, G.~Cui, and B.~Rigg, ``The development of the {CIE} 2000 colour-difference formula: {CIEDE2000},'' \emph{Color Research \& Application}, vol.~26, no.~5, pp. 340--350, 2001.

\bibitem{zhao2020towards}
Z.~Zhao, Z.~Liu, and M.~Larson, ``Towards large yet imperceptible adversarial image perturbations with perceptual color distance,'' in \emph{CVPR}, 2020, pp. 1036--1045.

\bibitem{Chattopadhyay2017GradCAMGG}
A.~Chattopadhyay, A.~Sarkar, P.~Howlader, and V.~N. Balasubramanian, ``Grad-cam++: Generalized gradient-based visual explanations for deep convolutional networks,'' \emph{WACV}, pp. 839--847, 2017.

\bibitem{Redmon2016YouOL}
J.~Redmon, S.~Divvala, R.~Girshick, and A.~Farhadi, ``You only look once: Unified, real-time object detection,'' in \emph{CVPR}, 2016.

\bibitem{He2016DeepRL}
K.~He, X.~Zhang, S.~Ren, and J.~Sun, ``Deep residual learning for image recognition,'' in \emph{CVPR}, 2016, pp. 770--778.

\bibitem{Simonyan2015VeryDC}
K.~Simonyan and A.~Zisserman, ``Very deep convolutional networks for large-scale image recognition,'' in \emph{ICLR}, 2015.

\bibitem{Szegedy2016RethinkingTI}
C.~Szegedy, V.~Vanhoucke, S.~Ioffe, J.~Shlens, and Z.~Wojna, ``Rethinking the inception architecture for computer vision,'' in \emph{CVPR}, 2016, pp. 2818--2826.

\bibitem{dosovitskiy2020}
A.~Dosovitskiy \emph{et~al.}, ``An image is worth 16x16 words: Transformers for image recognition at scale,'' in \emph{ICLR}, 2021.

\end{thebibliography}


\begin{thebibliography}{1}

\bibitem{dosovitskiy2020}
A. Dosovitskiy et al., ``An Image is Worth 16x16 Words: Transformers for Image Recognition at Scale,'' in \textit{ICLR}, 2021.

\bibitem{Liu2022AConvNet} 
Z. Liu, H. Mao, C.-Y. Wu, C. Feichtenhofer, T. Darrell, and S. Xie, ``A ConvNet for the 2020s,'' in \textit{CVPR}, 2022, pp. 11966-11976.

\bibitem{Tan2019efficientNet} 
M. Tan and Q. V. Le, ``EfficientNet: Rethinking model scaling for convolutional neural networks,'' in \textit{ICML}, 2019, pp. 10691-10700.



\bibitem{Howard2019SearchingFM} 
A. Howard, M. Sandler, G. Chu, L.-C. Chen, B. Chen, M. Tan, W. Wang, Y. Zhu, R. Pang, V. Vasudevan, Q. V. Le, and H. Adam, ``Searching for MobileNetV3,'' in \textit{ICCV}, 2019, pp. 1314–1324.

\bibitem{Liu2021SwinTransformer} 
Z. Liu, Y. Lin, Y. Cao, H. Hu, Y. Wei, Z. Zhang, S. Lin, and B. Guo, ``Swin Transformer: Hierarchical vision transformer using shifted windows,'' in \textit{ICCV}, 2021, pp. 9992-10002.

\end{thebibliography}
\clearpage 

\supptitle{CAPAA: Classifier-Agnostic Projector-Based Adversarial Attack}

\makeMySupplementaryTitle

\def\eg{\emph{e.g}.} \def\Eg{\emph{E.g}.}
\def\ie{\emph{i.e}.} \def\Ie{\emph{I.e}.}
\def\cf{\emph{c.f}.} \def\Cf{\emph{C.f}.}
\def\etc{\emph{etc}.} \def\vs{\emph{vs}.}
\def\wrt{w.r.t.} \def\dof{d.o.f.}
\def\etal{\emph{et al}.}

\renewcommand{\figureautorefname}{Fig.}
\renewcommand{\tableautorefname}{Table} 
\renewcommand{\equationautorefname}{Eqn.} 
\renewcommand{\sectionautorefname}{\S} 
\renewcommand{\subsectionautorefname}{\S} 

\def\BibTeX{{\rm B\kern-.05em{\sc i\kern-.025em b}\kern-.08em
    T\kern-.1667em\lower.7ex\hbox{E}\kern-.125emX}}

\section{Introduction}

In this supplementary material, we present the results of adversarial attacks against Vision Transformers (ViTs)\cite{dosovitskiy2020}. Using \textbf{teddy} as the target object, we employ Grad-CAM to analyze attention maps and evaluate attack effectiveness. As demonstrated in Table~\ref{tab:table 1}, our proposed CAPAA method significantly outperforms SPAA in classifier-agnostic multi-pose untargeted attacks, achieving a 3× higher average attack success rate against ViT-Base-16. Additionally, Table~\ref{tab:table 2} reveals near-perfect success rates (93.75\%) under the original pose configuration. While these results demonstrate strong performance, enhancing transferability to newer ViT variants represents an interesting direction for future research.

\begin{table}[htbp]

\centering
\caption{ 
Quantitative comparisons for classifier-agnostic \textbf{multi-pose} \textbf{untargeted} attacks.}
\label{tab:table 1}
\resizebox{0.5\textwidth}{!}{
\begin{tabular}{@{}cclllllcc@{}}
	\toprule[0.5mm]

\multicolumn{1}{l}{\textbf{Attacker}} & \multicolumn{1}{l}{\textbf{d$_\textbf{thr}$}} & \textbf{Classifier} & \textbf{SSIM$\uparrow$} & \textbf{L$_\textbf{2}\downarrow$} & \textbf{$\Delta \textbf{E}\downarrow$} & \textbf{L$_\textbf{inf}\downarrow$} & \textbf{U.top-1} & \multicolumn{1}{l}{\textbf{\begin{tabular}[c]{c@{}}\textbf{Avg. attack} \\ \textbf{success rate}\end{tabular}}} \\ \midrule
\multirow{16}{*}{CAPAA}               & \multirow{4}{*}{2}                  & Inception v3       & \multicolumn{1}{c}{\multirow{4}{*}{\textbf{0.902}}} & \multicolumn{1}{c}{\multirow{4}{*}{14.13}} & \multicolumn{1}{c}{\multirow{4}{*}{3.75}} & \multicolumn{1}{c}{\multirow{4}{*}{10.27}} & 20\%                    & \multirow{16}{*}{\textbf{48.75\%}}              \\
                                      &                                     & Resnet-18            & \multicolumn{1}{c}{}                       & \multicolumn{1}{c}{}                       & \multicolumn{1}{c}{}                      & \multicolumn{1}{c}{}                       & 20\%                    &                                                 \\
                                      &                                     & VGG-16               & \multicolumn{1}{c}{}                       & \multicolumn{1}{c}{}                       & \multicolumn{1}{c}{}                      & \multicolumn{1}{c}{}                       & 40\%                    &                                                 \\
                                      &                                     & ViT-Base-16          & \multicolumn{1}{c}{}                       & \multicolumn{1}{c}{}                       & \multicolumn{1}{c}{}                      & \multicolumn{1}{c}{}                       & 0                      &                                                 \\ \cmidrule(lr){2-8}
                                      & \multirow{4}{*}{3}                  & Inception v3       & \multicolumn{1}{c}{\multirow{4}{*}{0.861}} & \multicolumn{1}{c}{\multirow{4}{*}{16.15}} & \multicolumn{1}{c}{\multirow{4}{*}{4.40}} & \multicolumn{1}{c}{\multirow{4}{*}{11.98}} & 20\%                    &                                                 \\
                                      &                                     & Resnet-18            & \multicolumn{1}{c}{}                       & \multicolumn{1}{c}{}                       & \multicolumn{1}{c}{}                      & \multicolumn{1}{c}{}                       & 40\%                    &                                                 \\
                                      &                                     & VGG-16               & \multicolumn{1}{c}{}                       & \multicolumn{1}{c}{}                       & \multicolumn{1}{c}{}                      & \multicolumn{1}{c}{}                       & \textbf{100\%}                     &                                                 \\
                                      &                                     & ViT-Base-16          & \multicolumn{1}{c}{}                       & \multicolumn{1}{c}{}                       & \multicolumn{1}{c}{}                      & \multicolumn{1}{c}{}                       & 20\%                    &                                                 \\ \cmidrule(lr){2-8}
                                      & \multirow{4}{*}{4}                  & Inception v3       & \multicolumn{1}{c}{\multirow{4}{*}{0.828}} & \multicolumn{1}{c}{\multirow{4}{*}{18.66}} & \multicolumn{1}{c}{\multirow{4}{*}{5.32}} & \multicolumn{1}{c}{\multirow{4}{*}{14.17}} & 20\%                    &                                                 \\
                                      &                                     & Resnet-18            & \multicolumn{1}{c}{}                       & \multicolumn{1}{c}{}                       & \multicolumn{1}{c}{}                      & \multicolumn{1}{c}{}                       & 80\%                   &                                                 \\
                                      &                                     & VGG-16               & \multicolumn{1}{c}{}                       & \multicolumn{1}{c}{}                       & \multicolumn{1}{c}{}                      & \multicolumn{1}{c}{}                       & \textbf{100\%}                     &                                                 \\
                                      &                                     & ViT-Base-16          & \multicolumn{1}{c}{}                       & \multicolumn{1}{c}{}                       & \multicolumn{1}{c}{}                      & \multicolumn{1}{c}{}                       & 20\%                    &                                                 \\ \cmidrule(lr){2-8}
                                      & \multirow{4}{*}{5}                  & Inception v3       & \multicolumn{1}{c}{\multirow{4}{*}{0.807}} & \multicolumn{1}{c}{\multirow{4}{*}{21.13}} & \multicolumn{1}{c}{\multirow{4}{*}{6.18}} & \multicolumn{1}{c}{\multirow{4}{*}{16.22}} & 80\%                   &                                                 \\
                                      &                                     & Resnet-18            & \multicolumn{1}{c}{}                       & \multicolumn{1}{c}{}                       & \multicolumn{1}{c}{}                      & \multicolumn{1}{c}{}                       & \textbf{100\%}                     &                                                 \\
                                      &                                     & VGG-16               & \multicolumn{1}{c}{}                       & \multicolumn{1}{c}{}                       & \multicolumn{1}{c}{}                      & \multicolumn{1}{c}{}                       & \textbf{100\%}                     &                                                 \\
                                      &                                     & ViT-Base-16          & \multicolumn{1}{c}{}                       & \multicolumn{1}{c}{}                       & \multicolumn{1}{c}{}                      & \multicolumn{1}{c}{}                       & 20\%                    &                                                 \\ \midrule
\multirow{16}{*}{SPAA}                & \multirow{4}{*}{2}                  & Inception v3       & 0.878                                      & 15.44                                      & 3.87                                      & 11.19                                      & 5\%                   & \multirow{16}{*}{15.31\%}              \\
                                      &                                     & Resnet-18            & 0.882                                      & 15.37                                      & 3.95                                      & 11.19                                      & 5\%                   &                                                 \\
                                      &                                     & VGG-16               & 0.869                                      & \textbf{13.85}                                      & 3.72                                      & 10.21                                      & 10\%                    &                                                 \\
                                      &                                     & ViT-Base-16          & 0.874                                      & 13.90                                      & \textbf{3.64}                                      & \textbf{10.20}                                      & 5\%                   &                                                 \\ \cmidrule(lr){2-8}
                                      & \multirow{4}{*}{3}                  & Inception v3       & 0.840                                      & 17.13                                      & 4.54                                      & 12.78                                      & 5\%                   &                                                 \\
                                      &                                     & Resnet-18            & 0.834                                      & 15.63                                      & 4.35                                      & 11.69                                      & 15\%                   &                                                 \\
                                      &                                     & VGG-16               & 0.807                                      & 16.47                                      & 4.56                                      & 12.38                                      & 25\%                   &                                                 \\
                                      &                                     & ViT-Base-16          & 0.835                                      & 15.65                                      & 4.34                                      & 11.76                                      & 5\%                   &                                                 \\ \cmidrule(lr){2-8}
                                      & \multirow{4}{*}{4}                  & Inception v3       & 0.812                                      & 18.15                                      & 5.25                                      & 13.92                                      & 5\%                   &                                                 \\
                                      &                                     & Resnet-18            & 0.805                                      & 18.09                                      & 5.24                                      & 13.89                                      & 25\%                   &                                                 \\
                                      &                                     & VGG-16               & 0.775                                      & 19.46                                      & 5.60                                      & 14.93                                      & 40\%                    &                                                 \\
                                      &                                     & ViT-Base-16          & 0.809                                      & 18.01                                      & 5.14                                      & 13.72                                      & 5\%                   &                                                 \\ \cmidrule(lr){2-8}
                                      & \multirow{4}{*}{5}                  & Inception v3       & 0.801                                      & 18.91                                      & 5.64                                      & 14.61                                      & 10\%                    &                                                 \\
                                      &                                     & Resnet-18            & 0.792                                      & 19.39                                      & 6.00                                      & 15.20                                      & 35\%                   &                                                 \\
                                      &                                     & VGG-16               & 0.758                                      & 21.73                                      & 6.48                                      & 16.90                                      & 45\%                   &                                                 \\
                                      &                                     & ViT-Base-16          & 0.795                                      & 19.62                                      & 6.07                                      & 15.29                                      & 5\%                   &                                                 \\ \bottomrule
\end{tabular}
}
\end{table}

\begin{table}[htbp]
\centering
\caption{ 
Quantitative comparisons for classifier-agnostic \textbf{pose-specific} \textbf{untargeted} attacks.}
\label{tab:table 2}
\resizebox{0.5\textwidth}{!}{ 

\begin{tabular}
{@{}lllllrc@{}}
	\toprule[0.5mm]
\textbf{Attacker}                                                                                                & 
\textbf{Classifier} & 
\textbf{$\Delta \textbf{E}\downarrow$} & \textbf{SSIM$\uparrow$} & \textbf{L$_\textbf{2}\downarrow$} & \textbf{U.top-1} & \textbf{\begin{tabular}[c]{c@{}}Avg. attack \\success rate\end{tabular}}     \\
\addlinespace[3pt]
\hline 
\addlinespace[3pt]
\multirow{4}{*}{\begin{tabular}[c]{@{}c@{}}CAPAA\\ (ours)\end{tabular}} & Inception v3       & \multicolumn{1}{l}{\multirow{4}{*}{4.92}} & \multicolumn{1}{l}{\multirow{4}{*}{0.838}} & \multicolumn{1}{l}{\multirow{4}{*}{17.86}} & \textbf{100\%} & \multirow{4}{*}{\textbf{93.75\%}}                                 \\ 
                                                                        & Resnet-18            & \multicolumn{1}{c}{}                      & \multicolumn{1}{c}{}                       & \multicolumn{1}{c}{}                       &\textbf{100\%}                        &                                                          \\
                                                                        & VGG-16               & \multicolumn{1}{c}{}                      & \multicolumn{1}{c}{}                       & \multicolumn{1}{c}{}                       &\textbf{100\%}                        &                                                          \\
                                                                        & ViT-Base-16          & \multicolumn{1}{c}{}                      & \multicolumn{1}{c}{}                       & \multicolumn{1}{c}{}                       &  75\%                      &                                                          \\ \addlinespace[1pt] \hline \addlinespace[2.5pt]
\multirow{4}{*}{SPAA}                                                   & Inception v3       & 4.85                                      & 0.804                                      & \textbf{17.18}                                      & 25\%                   & \multirow{4}{*}{35.94\%}                                 \\
                                                                        & Resnet-18            & 5.07                                      & 0.798                                      & 17.87                                      & 44\%                   &                                                          \\
                                                                        & VGG-16               & 5.20                                      & \textbf{0.773}                                      & 18.27                                      & 50\%                   &                                                          \\
                                                                        & ViT-Base-16          & \textbf{4.86}                                      & 0.801                                      & 16.82                                      & 25\%                   &                                                          \\
                                                                        \bottomrule[0.3mm]
\end{tabular}
}
\end{table}

We also evaluated our method through comprehensive adversarial attacks across ten distinct experimental setups, each comprising 10 objects with 7 poses per object (totaling 70 test cases per setup). The evaluation covered four unseen classifier architectures: ConvNeXt-Base\cite{Liu2022AConvNet}, EfficientNet-B0\cite{Tan2019efficientNet}, MobileNetV3-Large\cite{Howard2019SearchingFM}, and Swin Transformer-Base\cite{Liu2021SwinTransformer}. As demonstrated in \autoref{tab:table 3}, our approach consistently outperforms the baseline across all classifiers and test conditions. While these results confirm the robustness of our method under varied pose-object combinations, we identify opportunities for further enhancement in cross-architecture transferability.

\begin{table}[htbp]
\centering
\caption{Average attack success rate for classifier-agnostic \textbf{multi-pose} \textbf{untargeted} attacks. ConvNeXt.B, MobileNetV3.L and Swin TF. B stand for ConvNeXt-Base, MobileNetV3 Large and Swin Transformer Base, respectively.}
\label{tab:table 3}
\resizebox{0.5\textwidth}{!}{
\begin{tabular}{@{\hspace{0pt}}l@{\hskip 6pt}c@{\hskip 5pt}c@{\hskip 5pt}c@{\hskip 5pt}c@{\hspace{0pt}}}

	\toprule[0.5mm]
\addlinespace[5pt] 
\textbf{Attacker} & \textbf{ConvNeXt.B} & \textbf{EfficientNet-B0} & \textbf{MobileNetV3.L} & \textbf{Swin TF. B} \\ \addlinespace[3pt] 
\hline
\addlinespace[5pt] 
SPAA              & 36.67\%               & 47.26\%                 & 54.64\%                   & 26.67\%             \\\addlinespace[5pt] 
\hline
\addlinespace[3pt]
\begin{tabular}[c]{@{}c@{}}CAPAA\\ (ours)\end{tabular}      & \textbf{38.57\%}      & \textbf{50.71\%}        & \textbf{58.21\%}          & \textbf{29.29\%}    \\ 
\bottomrule[0.3mm]
\end{tabular}
}
\end{table}

\end{document}